\documentclass{article} 
\usepackage{array}
\usepackage{iclr2024_conference,times}
\usepackage{multirow}
\usepackage{graphicx}

\usepackage{amsmath,amsfonts,bm}

\def\eqref#1{equation~\ref{#1}}

\def\1{\bm{1}}

\DeclareMathAlphabet{\mathsfit}{\encodingdefault}{\sfdefault}{m}{sl}
\SetMathAlphabet{\mathsfit}{bold}{\encodingdefault}{\sfdefault}{bx}{n}

\usepackage{tabularx} 

\usepackage{hyperref}
\usepackage{url}

\usepackage{booktabs}

\title{VidGen-1M: A Large-Scale Dataset for Text-to-video Generation}
\author{%
  Zhiyu Tan$^1$ \quad
  Xiaomeng Yang$^2$ \quad
  Luozheng Qin$^2$ \quad
  Hao Li\thanks{Corresponding Author.}~~$^1$ \\ [8pt]
  $^1$ Fudan University \quad 
  $^2$ Shanghai Academy of AI for Science \\ [8pt] \url{https://sais-fuxi.github.io/projects/vidgen-1m} \\
}

\iclrfinalcopy 
\begin{document}

\maketitle

\vspace{-6mm}
\begin{figure*}[ht]
\begin{center}
\includegraphics[width = 0.8\linewidth]{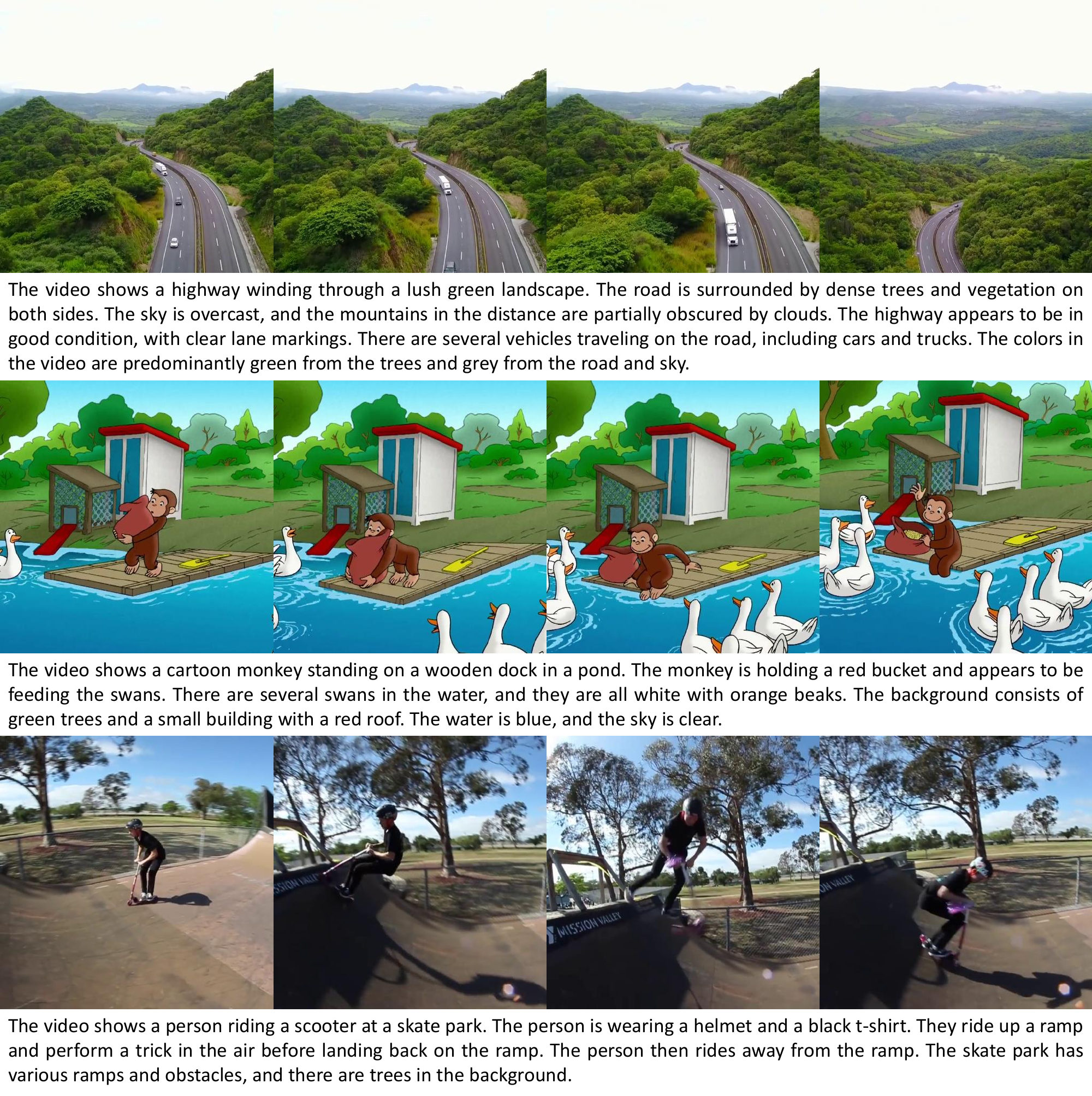}
\end{center}
\setlength{\abovecaptionskip}{0.cm}
\caption{\small A snapshot of the video-text pairs in VidGen-1M.
}
\label{fig:teaser}
\end{figure*}
\vspace{-1.5mm}
\begin{abstract}
The quality of video-text pairs fundamentally determines the upper bound of text-to-video models. Currently, the datasets used for training these models suffer from significant shortcomings, including low temporal consistency, poor-quality captions, substandard video quality, and imbalanced data distribution. The prevailing video curation process, which depends on image models for tagging and manual rule-based curation, leads to a high computational load and leaves behind unclean data. As a result, there is a lack of appropriate training datasets for text-to-video models. To address this problem, we present VidGen-1M, a superior training dataset for text-to-video models. Produced through a coarse-to-fine curation strategy, this dataset guarantees high-quality videos and detailed captions with excellent temporal consistency. When used to train the video generation model, this dataset has led to experimental results that surpass those obtained with other models.

\end{abstract}
\section{Introduction}

Recently, there have been significant advancements in text-to-video models, such as Latte~\citep{ma2024latte}, SORA, OpenSora~\citep{opensora}, and Window Attention Latent Transformer (W.A.L.T)~\citep{gupta2023walt}. However, training on current video-text datasets faces challenges, including unstable training and poor performance. These issues stem from several problems inherent in existing video-text datasets:
1)Low Quality Captions: The captions exhibit poor consistency with the videos and lack detailed captions. As demonstrated in DALLE-3~\citep{betker2023dalle3}, training text-to-image models on descriptive synthetic captions (DSC) significantly improves the performance on text-to-image alignment, which is also called "prompt following". While existing text-to-video datasets, such as HD-VILA-100M~\citep{xue2022hdvila} and Pandas-70M~\citep{chen2024panda}, particularly lack precise and detailed captions, the average length of captions in these datasets is less than 15 words. These captions often fail to capture object motion, actions, and camera movements, hindering the model's ability to effectively learn the semantic and temporal information in the videos.
2)Low Quality Videos: The existing dataset suffers from poor video quality and aesthetics, resulting in models trained on this dataset being unable to generate high-quality videos.
3)Temporal Inconsistency: The scene splitting algorithm fails to accurately detect scene transitions within videos, leading to instability in training models based on such data.
4)Data Imbalance: Current datasets, primarily composed of videos sourced from the internet, are often dominated by indoor human scenes, leading to significant data imbalance issues.

There are two main challenges in curating a dataset suitable for text-to-video generation tasks.
First, the existing curation process relies on either image models or optical flow models. Image models, such as the CLIP~\citep{radford2021clip}, lack the capability to capture temporal relationships in videos. Conversely, using optical flow scores to curate videos with fast camera movements and static scenes is inaccurate.
Second, compared to the data curation process for image-text pairs, the process for video-text pairs is significantly more complex and computationally intensive, posing major challenges for the academic community.

To tackle the aforementioned challenges, we introduce a multi-stage data curation process composed of three stages: coarse curation, captioning, and fine curation.
During the coarse curation stage, we utilize existing models to perform scene splitting and tagging on the videos. Based on these tags, we filter and sample the videos to create a curated dataset. This process ensures a balanced distribution across different categories and styles while reducing the number of videos that need to be processed in the subsequent computationally intensive curating phase. 
During the captioning stage, we employ a video captioning model to generate descriptive synthetic captions (DSC) for the videos.
In the fine curation stage, we employ a large language model (LLM) to refine video captions. This process addresses errors from the coarse curation stage, such as incorrectly filtered videos (e.g., those with scene transitions) and errors in caption generation (e.g., captions missing the EOS token).

In this work, we introduce a large-scale dataset comprising 1 million video clips with descriptive synthetic captions (DSC). The dataset features high-quality, open-domain videos accompanied by rich captions averaging 89.2 words each. These captions not only ensure stronger text-video alignment but also accurately capture the dynamic elements of the videos. Furthermore, the improved temporal consistency of the videos mitigates instability during model training. Additionally, the distribution of videos across different categories and styles is balanced. Training text-to-video models on our proposed dataset achieves superior performance compared to existing methods.

In summary, our main contributions can be summarized as follows:
\begin{itemize}

    \item We introduce a high-quality video dataset specifically designed for training text-to-video models.
    \item We propose a multi-stage curation method that achieves precise, high-quality curated data with limited computational resources.
    \item We release our text-to-video model, which generates high-quality videos that surpass the performance of existing state-of-the-art methods.
\end{itemize}

\section{Related work}
\begin{table}[t]
    \centering
    \small
    \label{tab:dataset}
  \resizebox{\textwidth}{!}{
  \begin{tabular}{l|cccccccc}
    \toprule
    \multicolumn{1}{c|}{Dataset} & Year & Text & Domain & \#Videos & AVL & ATL & Res \\
    \midrule
    MSVD~\citep{chen2011msvd}               & 2011 & Human    & Open    & 2K   & 9.7s  & 8.7 words  & -     \\
    LSMDC~\citep{rohrbach2015lsmdc}                     & 2015 & Human    & Movie   & 118K   & 4.8s  & 7.0 words  & 1080p \\
    UCF101~\citep{soomro2012ucf101}         & 2015 & Human   & Action   & 13K    & 7.2s     & 4.3 words & 240p \\
    MSR-VTT~\citep{xu2016msrvtt}            & 2016 & Human    & Open    & 10K    & 15.0s & 9.3 words  & 240p  \\
    DiDeMo~\citep{anne2017ldidemo}          & 2017 & Human    & Flickr  & 27K    & 6.9s  & 8.0 words  & -     \\
    ActivityNet~\citep{caba2015activitynet}         & 2017 & Human    & Action  & 100K   & 36.0s & 13.5 words & -     \\
    YouCook2~\citep{zhou2018youcook2}               & 2018 & Human    & Cooking & 14K    & 19.6s & 8.8 words  & -     \\
    VATEX~\citep{wang2019vatex}             & 2019 & Human    & Open    & 41K    & $\sim$10s & 15.2 words & - \\
    HowTo100M~\citep{miech2019howto100m}    & 2019 & ASR               & Open    & 136M   & 3.6s  & 4.0 words  & 240p  \\
    YT-Temporal-180M~\citep{zellers2021yt180m} & 2021 & ASR            & Open    & 180M   & -     & -          & -     \\
    HD-VILA-100M~\citep{xue2022hdvila}     & 2022 & ASR                & Open    & 103M   & 13.4s & 32.5 words & 720p  \\
    Panda-70M~\citep{chen2024panda}                             & 2024 & Auto & Open    & 70.8M  & 8.5s  & 13.2 words & 720p \\
    \textbf{VIDGEN-1M~(Ours)}                          & 2024 & Auto & Open    & 1M     & 10.6s & 89.3 words          & 720p \\
    \bottomrule
    \end{tabular}}
    \caption{\small Comparison of our dataset and other video-text datasets. ``AVL'' and ``ATL'' are abbreviations for ``Average Video Length'' and ``Average Text Length'', respectively.} 
    \vspace{-3.5mm}
\end{table}
\subsection{Video-text Dataset}
To facilitate the development of video understanding and generation, researchers build a large volume of video-text datasets that vary in video length, resolution, domain, and scale.
For instance, UCF101~\citep{soomro2012ucf101} is originally an action recognition dataset consisting of 13,320 videos, which can be classified into 101 categories.
Formulate a unified text conditions for each category, UCF101 is widely used for benchmarking text-to-video generation.
MSVD~\citep{chen2011msvd} and MSRVTT~\citep{xu2016msrvtt} are two open-domain video-text datasets popular in video retrieval.
These datasets collect videos first and then annotate these videos with human annotators.
However, due to the heavy cost of human annotation, they are usually limited by scale, usually at thousands scale.
To alleviate this and expand video-text datasets to million scale, How2100M, HD-VILA-100M~\citep{xue2022hdvila} and YT-Temporal-180M~\citep{zellers2021yt180m} propose to automatically annotate videos with subtitles generated by ASR models.
Meanwhile, Webvid scrapes 10.7 million videos along with text annotation.
While Panda-70M~\cite{chen2024panda} collects 70 million high-resolution and semantically coherent video samples.

These large scale video-text datasets surly lay the cornerstone for the adavancement of text-to-video generation.
However, they are limited by low quality captions, low video quality, temporal inconsistency and data imbalance.
To alleviate these challenges, we meticulously curate Panda-70M in a coarse-to-fine way. 
Owing to our comprehensive and effective data curation, VidGen-1M features high video quality, high video-text consistency and balanced and diverse video content, which significantly differ it with previous works.

\subsection{Text-to-video generation model}
To investigate the best practice for designing video generation models, researchers has made a series of progress.
SVD~\citep{blattmann2023svd} first utilize SDXL~\citep{sdxl} to generate images conditioning on the input text, and then generate videos based on the generated images.
MAGVIT2~\citep{yu2023magvit2} is a VQGAN~\citep{esser2021vqgan} model that addresses the problem of codebook size and utilization by employing the lookup-free technique and training a large codebook.
Specifically, it maps videos into quantized video token sequences and generates videos in an autoregressive manner.
WALT~\citep{gupta2023walt} proposes to patchify input videos to lower the training costs.
Meanwhile, Latte~\citep{ma2024latte} initializes its parameters from Pixart~\citep{chenpixart}, and investigate the training effciency of 4 DiT variants. 
SORA sparks the revolution of text-to-video generation, emerging a series of DiT~\citep{peebles2023dit}-based video generation models, such as OpenSora~\citep{opensora} and Mira~\citep{ju2024mira}.
Except for these open source models, there are also some commercial video generation models that exhibit strong generation performance, such as kling, dreamachine, and vidu.

\section{Method}
In the construction of VidGen, we harnessed 3.8 million high-resolution, long-duration videos derived from the HD-VILA dataset. These videos were subsequently split into 108 million video clips. Following this, we tagged and sampled these video clips. The VILA model was then utilized for video captioning. Lastly, to rectify any data curating errors from the preceding steps, we deployed the LLM for further caption curating.




\subsection{Coarse Curation}
To achieve efficient data curation with limited computational resources, we first employ a coarse curation approach. This involves scene splitting, video tagging, filtering, and sampling to reduce the computational load in subsequent stages of captioning and fine curation.

\subsubsection{Scene Splitting}

Motion inconsistencies, such as scene changes and fades, are frequently observed in raw videos.
However, since motion inconsistencies directly cut off the video semantics, text-to-video models are significantly sensitive to and confused by them, leading to heavy impairment on training efficiency.
To alleviate their impairment, we follow the prior research~\citep{blattmann2023svd, chen2024panda,opensora} to utilize PySceneDetect~\citep{pyscenedetect} in a cascading manner to detect and remove scene transitions in our raw videos. 

\subsubsection{Tagging}
In order to construct a dataset suitable for training text-to-video models, the data must meet the following criteria: high-quality videos, balanced categories, and strong temporal consistency within the videos. To achieve this goal, we first need to tag each splitted video clip. Subsequently, these tags serve as the basis for curating and sampling.

\textbf{Video Quality.}
The visual quality of videos is of paramount importance for the efficient training of text-to-video models. 
In order to enhance the quality of generated videos in text-to-video generation, we adopt a strategy of filtering out videos with low aesthetic appeal and high OCR scores. 
In this context, we employ the LAION Aesthetics model to predict and evaluate aesthetic scores, thereby ensuring a superior quality of training data.
Particularly, the aesthetics models can also filter out visually abnormal videos, such as videos with irregular color distribution or weird visual elements.

\textbf{Temporal Consistency.} 
Incorrect scene splitting in videos can significantly impair the effectiveness of model training. High temporal consistency is a crucial characteristic required for the training data in text-to-video models. To ensure this, we utilize the CLIP model to extract visual features and assess temporal consistency. This assessment is achieved by calculating the cosine similarity between the starting and ending frames of video clips, thereby providing a quantitative measure of continuity and coherence.

\textbf{Category} 
The HD-VILA-100M video dataset displays significant imbalances across its categories, resulting in less than optimal performance of video generation models for these categories. To tackle this issue, we deploy predefined category tags to label each video, with the assistance of the CLIP model. Specifically, we extract the CLIP image features from the initial, middle, and final frames of each video, compute their average, and then determine the similarity between these averaged image features and the textual features associated with each tag. This methodology enables us to assign the most fitting tags to each video.

\textbf{Motion.} 
We employ the RAFT~\citep{teed2020raft} model to predict the optical flow score of videos. As both static videos and those with excessively fast motion are detrimental for training text-to-video models, we filter out these videos based on their optical flow scores.

\subsubsection{Sampling}

\begin{figure}[ht]
    \centering
    \includegraphics[width=\textwidth]{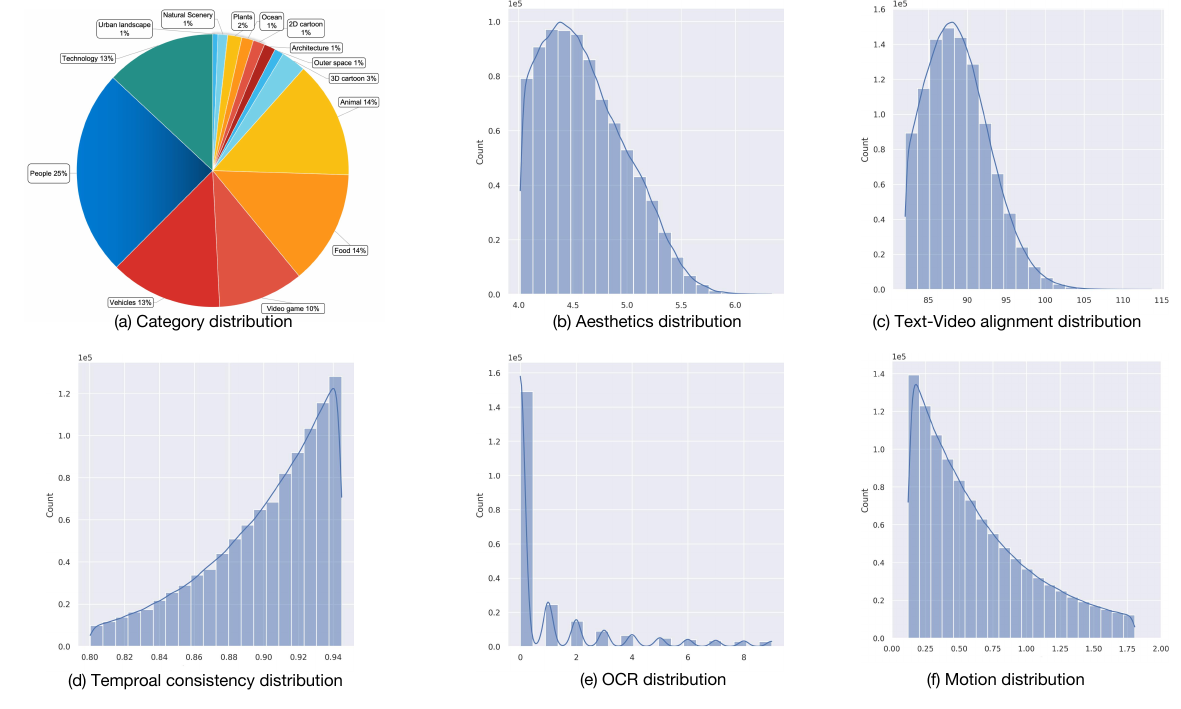}
    \caption{Distribution of curated data.}
    \label{fig:data_distribution}
\end{figure}

By employing tags associated with visual quality, temporal consistency, category, and motion, we undertook the task of filtering and sampling. The curated data distribution across multiple dimensions in our dataset is depicted in Figure \ref{fig:data_distribution}. This figure clearly indicates that videos characterized by low quality, static scene, excessive motion speed, and those demonstrating inadequate alignment between text and video along with poor temporal consistency were systematically removed. Concurrently, we have ensured a relatively even distribution of samples across diverse categories.

\subsection{Captioning}

\begin{table}[ht]
\centering
\begin{tabular}{l|cccc}
\toprule
\textbf{Dataset}      & \textbf{VN/DN}    & \textbf{VV/DV}    & \textbf{Avg N} & \textbf{Avg V} \\
\midrule
Pandas-70M            & 16.1\%            & 19.2\%            &   4.3             &   1.9      \\
\textbf{Ours}                  & 20.3\%            & 41.1\%            &   22.5             &   15.9     \\
\bottomrule
\end{tabular}
\caption{Statistics of noun and verb concepts for different datasets. 
VN: valid distinct nouns (appearing more than 10 times); DN: total distinct nouns; Avg N: average noun count per video.
VV: valid distinct verbs (appearing more than 10 times); DV: total distinct verbs; Avg N: average verbs count per video.}
\label{tab:dataset_summary}
\end{table}

The quality of video captions exerts a critical influence on text-to-video model, while the captions in the HD-VILA-100M dataset demonstrate several problems, including misalignment between text and video, inadequate descriptions, and limited vocabulary use. To enhance the information density of the captions, we employ the cutting-edge vision-language model, VILA~\citep{lin2024vila}. 
Owing to the remarkable video captioning ability of VILA, we have significantly enhanced caption quality. 
After captioning, we apply clip score to filter out the text-video pairs with low similarity.

We present a vocabulary analysis in Table \ref{tab:dataset_summary}, where we identify valid distinct nouns and valid distinct verbs as those that appear more than 10 times in the dataset. Utilizing the VILA model on the HD-VILA-100M dataset, we have generated the enhanced HD-VILA-100M dataset. In the Panda-70M dataset, there are 270K distinct nouns and 76K distinct verbs; however, only 16.1\% and 19.2\% of these meet the validity criteria, respectively. 
Captions generated using VILA substantially enhance the valid ratio as well as the average count of nouns and verbs per video, thereby increasing the conceptual density.

\subsection{Fine Curation}

\begin{figure}[ht]
    \centering
    \includegraphics[scale=0.5]{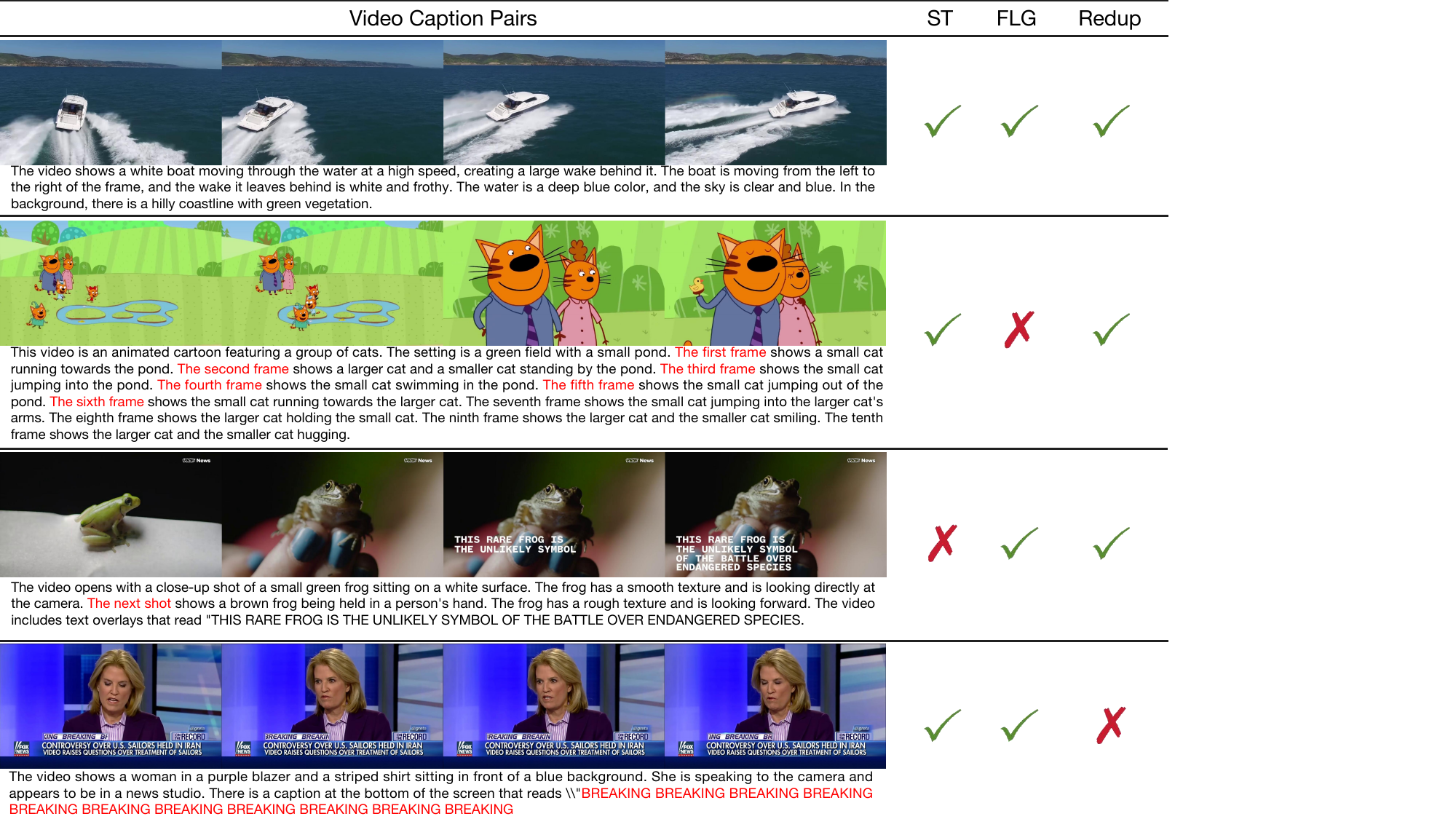}
    \caption{Caption Curation Result. Employing Llama3.1 to curate video captions, as produced by VILA, results in a significant improvement in the quality of the training dataset. This is specifically evidenced through improvements in video temporal consistency and alignment between text and video. Consequently, this method facilitates notable improvements in the performance of the text-to-video model.}
    \label{fig:caption_compare}
\end{figure}

In the stages of coarse curation and captioning, filtering for text-image alignment and temporal consistency using the CLIP score can remove some inconsistent data, but it is not entirely effective. Consequently, issues such as scene transitions in video, and two typical description errors occur in video captions: 1) Failed generating eos token, where the model fails to properly terminate the generation process, leading to looping or repetitive token generation, and 2) Frame-level generation, where the model lacks understanding of the dynamic relationships between frames and generates isolated descriptions for each frame, resulting in captions that lack coherence and fail to accurately reflect the video's overall storyline and action sequence.

To address the mentioned data curating issues, one potential solution is manual annotation, but this approach is prohibitively expensive. 
Fortunately, with recent advancements in large language models, this problem can be resolved.
Errors in captions generated by Multi-Modal Language Models (MLLMs) can be identified by analyzing specific patterns, such as scene transitions, repetitive content, and frame-level descriptions, using a Language Model (LLM).
Models like LLAMA3 have shown exceptional proficiency in these tasks, making them a viable alternative to manual annotation.

In our endeavor to isolate and remove video-text pairings that exhibit discrepancies in both text-video alignment and temporal consistency, we leveraged the cutting-edge Language Model (LLM), LLAMA3.1, to scrutinize the respective captions. The application of the fine curation has facilitated a marked improvement in the quality of the text-video pairs, as evidenced in Figure \ref{fig:caption_compare}. Our study primarily centers around three critical factors: Scene Transition (ST), Frame-level Generation (FLG), and Reduplication (Redup).

\section{Experiments}
\subsection{Implementation details.} 
\textbf{Experiment setup.}
To evaluate the effectiveness of our text-to-video training dataset, we performed a comprehensive evaluation using the base model, composed of both spatial and temporal attention blocks. To accelerate the training process, we initially conducted extensive pre-training on a large collection of low-resolution $256 \times 256$ images and videos. Following this, we carried out joint training on our VidGen-1M dataset using $512 \times 512$ px resolution images and 4-second videos.


\subsection{Experiment Results}

\begin{figure}[ht]
    \centering
    \includegraphics[width=\textwidth]{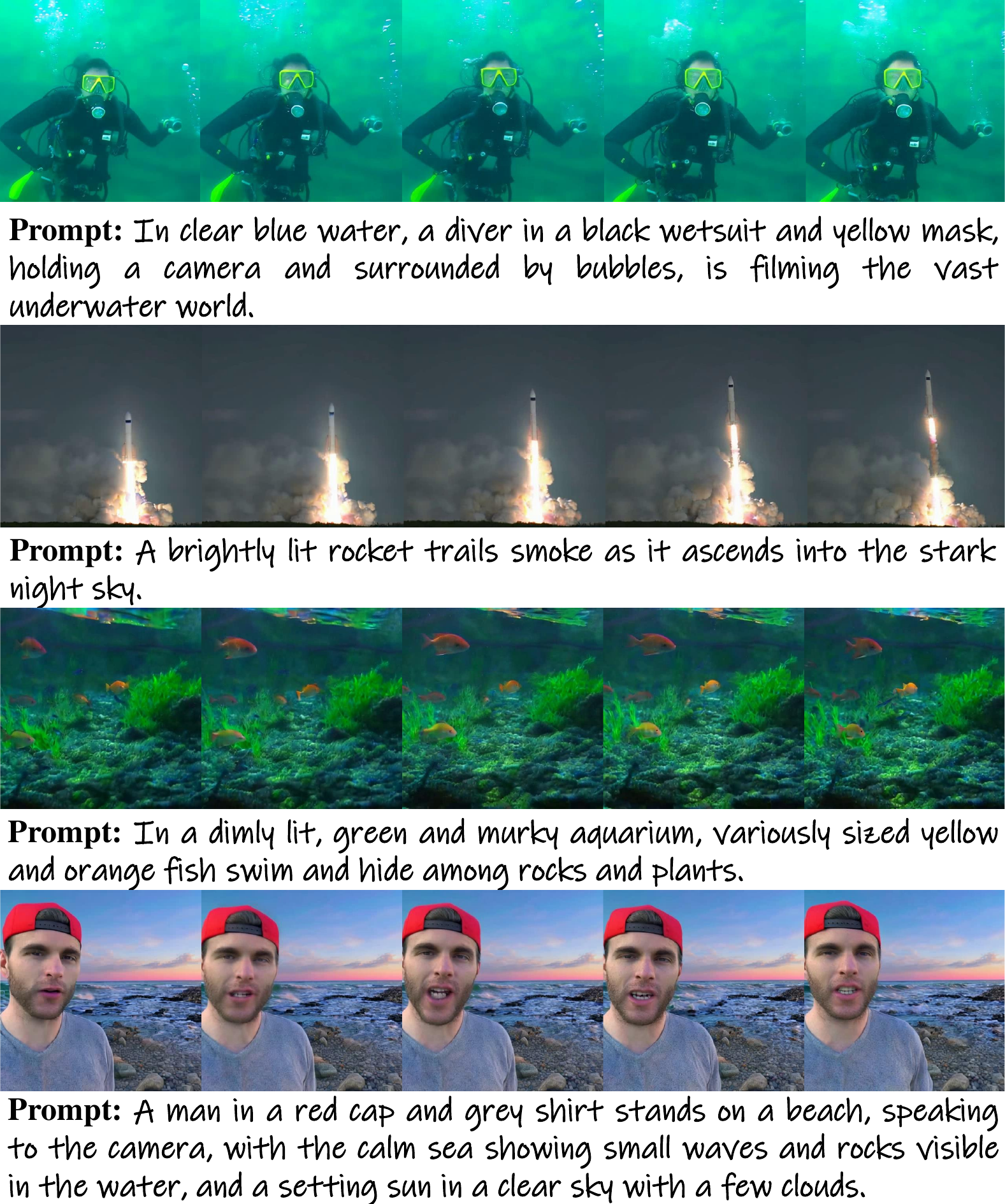}
    \caption{\textbf{Qualitative Evaluation}: The model we've developed can generate videos from natural language prompts at a resolution of 512 × 512. These videos are 4 seconds long and play at 8 frames per second. Notably, our model can generate photorealistic videos that maintain temporal consistency and align accurately with the textual prompt.}
    \label{fig:gen_vis}
\end{figure}



\subsubsection{Qualitative Evaluation.}

As displayed in Figure\ref{fig:gen_vis}, our model's ability to generate superior-quality videos is a testament to the robustness of the high-resolution VidGen-1M dataset. This dataset's high quality is reflected in the realism and detail of the generated videos, reinforcing its effectiveness in training our model. A noteworthy characteristic of our generated videos is their strong "prompt following" ability, which is a direct outcome of the high consistency between video-text pairs in the training data. This consistency ensures that the model can accurately interpret the textual prompts and generate corresponding video content with high fidelity. The first example further underlines the high quality of the VidGen-1M dataset. The generated video demonstrates remarkable realism - from the diver's hair floating underwater to the motion of the bubbles. These details, which showcase significant temporal consistency and adhere to real-world physics, highlight the model's capability to generate believable and visually accurate video content.

The VidGen-1M dataset's quality has far-reaching implications for the field of computer vision, particularly for text-to-video generation. By providing high-resolution and temporal consistency training data, VidGen-1M enables models to generate more realistic and high-quality videos. This can lead to advancements in video generation techniques, pushing the boundaries of what is currently possible.
Furthermore, the high-quality data provided by VidGen-1M could potentially streamline the model training process. With more accurate and detailed training data, models can learn more effectively, potentially reducing the need for extensive computational resources and time-consuming training periods. In this way, VidGen-1M not only improves the outcomes of text-to-video generation but also contributes to more efficient and sustainable model training practices.


\section{Conclusion}
In this paper, we introduce a high-quality video-text dataset features high video quality, high caption quality, high temporal consistency and high video-text alignment, specifically designed for the training of text-to-video generation models.
The aforementioned various high quality features arise from our meticulously data curation procedure, which efficiently ensures high data quality in a coarse-to-fine manner.
To verify the effectiveness of VidGen-1M, we train a text-to-video generation model on it.
The results are promising, the model trained on VidGen-1M achieves remarkably better FVD scores on zero-shot UCF101, compared with state-of-the-art text-to-video models.
To bootstrap the development of high performance video generation models, we will release VidGen-1M, along with the related codes and the models trained on it, to the public.

\bibliography{iclr2024_conference}
\bibliographystyle{iclr2024_conference}

\end{document}